\documentclass[10pt,twocolumn,letterpaper]{article}

\PassOptionsToPackage{table}{xcolor}

\usepackage{cvpr}              %

\definecolor{cvprblue}{rgb}{0.21,0.49,0.74}
\usepackage[pagebackref,breaklinks,colorlinks,allcolors=cvprblue]{hyperref}

\usepackage{graphicx}
\usepackage{amsmath}
\usepackage{amssymb}
\usepackage[scr=boondoxo]{mathalfa}
\usepackage{booktabs}
\usepackage{multirow}

\usepackage{adjustbox}

\newcommand{\Z}{\mathbb{Z}}

\newcommand\dotprod[1]{\left\langle#1\right\rangle}

\usepackage{soul}
\usepackage{booktabs}
\usepackage{dsfont}
\usepackage[numbers]{natbib}

\newtheorem{theorem}{Thrm}%

\def\paperID{31181} %
\def\confName{CVPR}
\def\confYear{2026}

\title{SASNet: Spatially-Adaptive Sinusoidal Networks for INRs}

\author{Haoan Feng\textsuperscript{1},
Diana Aldana\textsuperscript{2},
Tiago Novello\textsuperscript{2},
Leila De Floriani\textsuperscript{1}
\\[.2cm]
\textsuperscript{1}\normalsize{University of Maryland},\quad
\textsuperscript{2} \normalsize{IMPA}
}

\begin{document}
\maketitle
\begin{abstract}
Sinusoidal neural networks (SIRENs) are powerful implicit neural representations (INRs) for low-dimensional signals in vision and graphics. By encoding input coordinates with sinusoidal functions, they enable high-frequency image and surface reconstruction. However, training SIRENs is often unstable and highly sensitive to frequency initialization: small frequencies produce overly smooth reconstructions in detailed regions, whereas large ones introduce spurious high-frequency components that manifest as noise in smooth areas such as image backgrounds. 
To address these challenges, we propose \textbf{SASNet}, a \textit{Spatially-Adaptive Sinusoidal Network} that couples a \textit{frozen frequency embedding layer}, which explicitly fixes the network's frequency support, with \textit{jointly learned spatial masks} that localize neuron influence across the domain. This pairing stabilizes optimization, sharpens edges, and suppresses noise in smooth areas. 
Experiments on 2D image and 3D volumetric data fitting as well as signed distance field (SDF) reconstruction benchmarks demonstrate that SASNet achieves faster convergence, superior reconstruction quality, and robust frequency localization--assigning low- and high-frequency neurons to smooth and detailed regions respectively--while maintaining parameter efficiency.
Code available at here\footnote{\url{https://github.com/Fengyee/SASNet_inr}}.
\end{abstract}

\section{Introduction}
Implicit neural representations (INRs) have emerged as powerful tools for modeling low-dimensional signals in computer vision and graphics~\cite{park2019deepsdf,mescheder2019occupancy,sitzmann2019scene}. 
They encode signals implicitly in the weights of neural networks by mapping coordinates directly to signal values. 
In particular, sinusoidal networks (SIRENs)~\cite{sitzmann2020implicit} employ sinusoidal activations, allowing them to capture fine details by modeling high frequencies. 
These properties make SIRENs particularly well-suited for tasks that require accurate high-frequency reconstruction, such as image fitting~\cite{reddy2021multi,kim2024arbitrary,liu2024finer, shi2024improved}, super-resolution~\cite{chen2023neurbf,saragadam2023wire}, signed distance field (SDF) modeling~\cite{mescheder2019occupancy,sitzmann2020implicit,feng2024implicitterrain,novello2022exploring,Novello2023Neural, schirmer2024geometric}, and morphing~\cite{schardong2024neural, novello2025,chen2025nivm}.
\begin{figure}[tbp]
    \centering
    \captionsetup{type=figure}
    \includegraphics[width=\columnwidth]{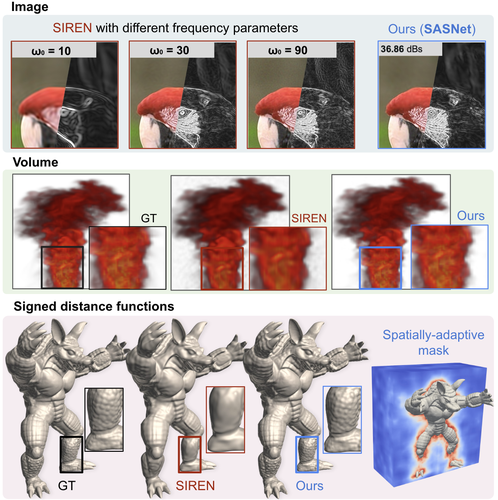}
    \vspace{-0.6cm}
    \caption{We present \textbf{SASNet}, a robust INR that introduces spatial localization into SIRENs, addressing challenges in frequency initialization, hyperparameter selection, and noisy reconstruction.
The first row compares SIREN~\cite{sitzmann2020implicit} with different frequency parameters~$\omega_0$ and SASNet on a~$\times16$ super-resolution task.
Low~$\omega_0$ values yield blurry reconstructions, while higher~$\omega_0$ values produce ringing artifacts and noisy backgrounds; in contrast, SASNet achieves sharper and cleaner results.
Similar results are observed in the 3D volumetric fitting task (second row).
Finally, SASNet achieves superior SDF fitting quality (Armadillo's leg) by learning spatially-adaptive masks that focus neuron activations near the zero-level set (bottom right).}\label{fig:teaser}\vspace{-0.2cm}
\end{figure}

Despite their expressive power, SIRENs are highly sensitive to hyperparameters and provide limited control over frequency support, often leading to reconstruction artifacts. Their performance depends critically on the input frequency parameter $\omega_0$, which determines the network's spectral range. As shown in~\cref{fig:teaser} (top), a small $\omega_0$ produces clean but overly smooth reconstructions, while a large $\omega_0$ improves sharpness but introduces high-frequency noise in smooth regions. We refer to this undesired activation of high-frequency components in low-frequency regions as \textit{frequency leakage}. Recovering high-frequency details by prolonged training often further destabilizes optimization and causes overfitting, degrading both generalization and derivative accuracy. These localization issues limit the practical use of SIRENs. Grid-based networks~\cite{takikawa2021neural,muller2022instant} alleviate this problem by interpolating spatial features with a hash grid followed by an MLP, but representing fine details still requires very high-resolution grids, increasing memory and computation costs and limiting scalability.

To address these issues, we propose \textbf{SASNet} (\textbf{S}patially-\textbf{A}daptive \textbf{S}inusoidal Neural \textbf{Net}work), an architecture that introduces robust and localized frequency control by making SIREN activations spatially dependent. 
SASNet adopts a frozen frequency embedding layer, following~\citet{novello2024taming}, to provide explicit control over the network's spectral support. 
To achieve frequency localization, SASNet modulates SIREN neuron activations by employing \textit{spatially-adaptive masks} generated from a lightweight hash-grid MLP. 
These masks are applied to each layer via element-wise multiplication, regulating the influence of different frequency bands and constraining neuron responses to specific spatial regions (see~\cref{fig:teaser}, second row). 
Both networks are trained jointly, reducing redundancy and overfitting while concentrating neuron contributions in unfitted regions.
\cref{fig:teaser} showcases SASNet's results on 2D image fitting and SDF reconstruction tasks, demonstrating significant improvements over prior methods in accuracy, speed, and spatial consistency.
Our main contributions are:
\begin{itemize}
    \item We introduce \textbf{SASNet}, which combines a dedicated \textbf{frequency embedding layer}~\cite{novello2024taming} with learned spatial modulation to bring localized frequency control to SIRENs.
    \item We propose a lightweight multi-scale hash-grid MLP that learns \textbf{spatially-adaptive masks} to localize the influence of different frequency bands and neuron groups, reducing high-frequency leakage, overfitting in smooth regions, and redundant neuron contributions.
    \item We show that this design yields more stable optimization and faster convergence than prior methods while improving reconstruction quality and spatial consistency across image fitting, volumetric reconstruction, and SDF~tasks.
\end{itemize}

\section{Related work}\label{sec:relatedwork}
\textbf{Implicit neural representations} (INRs) have become fundamental in computer vision and graphics for modeling low-dimensional signals~\cite{park2019deepsdf,mescheder2019occupancy,sitzmann2019scene, rezaeian2025sl2a, huang2025few, ben2024neural, walker2025spatially,mehta2021modulated,morsali2025staf,kim2025grids}. 
They encode spatial coordinates directly through neural network parameters, enabling differentiable and resolution-independent signal reconstruction. 
XINC~\cite{padmanabhan2024explaining} introduces neuron contribution maps to visualize how individual neurons in INRs contribute to the output. Adopting this view, we propose a three-category taxonomy that classifies INRs into \textit{global}, \textit{local}, and \textit{hybrid} representations.

\vspace{0.1cm}
In \textbf{global INRs}, each neuron contributes to the entire spatial domain. 
Early \texttt{ReLU}-based MLPs exhibit a strong spectral bias~\cite{Rahaman2018OnTS}, favoring low frequencies and struggling to capture high-frequency details~\cite{Rahaman2018OnTS,ramasinghe2022frequency}. 
To alleviate this limitation, FFN~\cite{tancik2020fourier} introduced positional encoding, while alternative activation functions such as sine~\cite{sitzmann2020implicit}, Gaussian~\cite{ramasinghe2022beyond}, and sinc~\cite{saratchandran2024sampling} have also been explored. 
SIREN~\cite{sitzmann2020implicit} is the most widely used INR due to its high representational capacity across diverse signal domains. 
However, its training is highly sensitive to the frequency parameter $\omega_0$. 
Recent work~\cite{Novello2023Neural} introduces improved initialization and weight-bounding strategies for explicit spectral control, while FINER~\cite{liu2024finer} proposes a dynamic scaling factor for adaptive frequency modulation. 
Other approaches mitigate spectral bias via weight reparameterization~\cite{shi2024improved} or extend frequency expressiveness using auxiliary techniques~\cite{mehta2021modulated,kazerouni2024incode}. 
Despite such advances, global INRs lack spatial localization, which reduces parameter efficiency and often produces blurry or noisy reconstructions (see~\cref{fig:teaser}).

\vspace{0.1cm}
\textbf{Local INRs} restrict neuron influence to specific spatial regions, improving scalability and localized detail modeling. 
Examples include sinc-activated networks~\cite{saratchandran2024sampling} and WIRE~\cite{saragadam2023wire}, which employ Gabor wavelets to encode spatial locality but require careful initialization and are prone to overfitting. 
Inspired by Gaussian splatting~\cite{kerbl3Dgaussians}, recent works~\cite{zhang2024gaussianimage,zhang2024image} use Gaussian kernels for compact image fitting, improving inference efficiency through parameter compression and gradient-based initialization. 
Although these models mitigate some limitations of global INRs, they lack explicit frequency control, often overfitting local textures, and typically require a larger parameter count.

\vspace{0.1cm}
\textbf{Hybrid INRs} integrate the strengths of global and local formulations by combining multi-scale features or decomposing the domain into sub-regions. 
InstantNGP~\cite{muller2022instant} employs multi-scale hash-grid encodings followed by an MLP for 3D scene representation, often requiring large grids and expensive training. 
MINER~\cite{saragadam2022miner} decomposes the domain into hierarchical tiles and iteratively fits residuals. 
NeuRBF~\cite{chen2023neurbf} merges fixed frequency embeddings with radial basis functions (RBFs) to improve local fitting, while SAPE~\cite{hertz2021sape} expands model capacity using residual-dependent masks that introduce high-frequency embeddings in poorly reconstructed regions. 
More recently, Walker~\etal~\cite{walker2025spatially} introduce a learned spatial gating mechanism that dynamically selects which levels of a multi-resolution hash grid contribute at each location, enabling spatially-adaptive frequency allocation that reduces artifacts in smooth regions.
Despite their advances, these methods can still overfit, show limited expressiveness under compact configurations, and degrade signal derivative accuracy due to \texttt{ReLU}-based activations.

\pagebreak
Our \textbf{SASNet} addresses these issues by combining the frequency control of SIRENs~\cite{novello2024taming} with hash-grid MLP's spatial modulation, resulting in a robust INR that achieves frequency localization, reduces overfitting in smooth regions, converges faster, and delivers superior reconstruction without increasing the number of~parameters.

\section{Method}\label{sec:method}

\begin{figure}[h!]
    \centering
    \includegraphics[width=\linewidth]{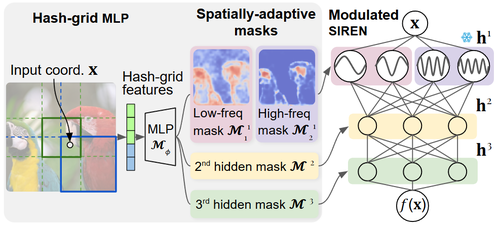}
    \vspace{-0.2cm} 
    \caption{
\textbf{Overview of SASNet.}
A hash-grid encodes the input coordinate $\mathbf{x}$ into features, which are decoded via a \texttt{ReLU} MLP into spatially-adaptive masks $\mathcal{M}^i(\mathbf{x})$.
Each mask modulates the corresponding layer of the sinusoidal MLP through an element-wise (Hadamard) product $\odot$, localizing neuron influence across the spatial domain.
The frozen frequency embedding layer $\mathbf{h}^1$ fixes the network spectrum, while the masks localize the neuron influence, stabilizing training and improving reconstruction.
}\label{fig:pipeline}
\vspace{-0.4cm}
\end{figure}

Our goal is to design a network capable of representing the fine details captured by SIRENs~\cite{sitzmann2020implicit} while avoiding the injection of unwanted high-frequencies into smooth regions of the signal. 
For this, we introduce the \textit{spatially-adaptive sinusoidal network} (SASNet), illustrated in~\cref{fig:pipeline}, which integrates a hash-grid MLP that modulates the neurons of a SIREN through learned adaptive masks, enabling localized frequency control. 
This allows preserving high-frequency detail in complex regions while maintaining smoothness in low-frequency areas, leading to a robust representation.

\subsection{Problem setting}
Let $\{\mathbf{x}_i, \mathscr{f}_i\}_{i=1}^N$ be samples of an unknown low-dimensional signal $\mathscr{f}$, we aim to learn a neural network $f_\theta$ that approximates $\mathscr{f}$ while avoiding overfitting. 
This setup follows the standard INR formulation, where the parameters $\theta$ are implicitly defined by the loss function:
\begin{align}
\mathscr{L}(\theta)
= \frac{1}{N} \sum_{i=1}^{N} \| f_{\theta}(\mathbf{x}_i) - \mathscr{f}_i \|_2^2
+ \lambda\, \mathscr{R}(\theta),
\end{align}
where the first term enforces data fitting, and the regularization term $\mathscr{R}(\theta)$ constrains $f_\theta$ to satisfy some property. 
For instance, for signed distance functions (SDFs)~\cite{gropp2020implicit}, $\mathscr{R}(\theta)$ enforces the Eikonal constraint $\lvert\nabla f_\theta\rvert \!=\! 1$. 

\vspace{0.1cm}
\noindent\textbf{Sinusoidal MLPs.}
SIREN~\cite{sitzmann2020implicit} is a common parameterization for $f_\theta$. It is defined as the composition of $d+1$ layers
\begin{equation}\label{siren}
f_\theta(\mathbf{x}) = \mathbf{L} \circ \mathbf{S}^d \circ \cdots \circ \mathbf{S}^0(\mathbf{x}),
\end{equation}
where each layer is given by $\mathbf{S}^i(\mathbf{x}) = \sin(\mathbf{W}^i \mathbf{x}\! + \mathbf{b}^i)$, parameterized by a weight matrix $\mathbf{W}^i \!\in\! \mathbb{R}^{n_{i+1} \times n_i}$ and bias $\mathbf{b}^i\!\in\! \mathbb{R}^{n_{i+1}}$. 
The first layer $\mathbf{S}^0$ maps the input coordinates $\mathbf{x}$ into sinusoidals $\sin(\omega \mathbf{x}\! +\! \varphi)$, where $\omega \!:=\! \mathbf{W}^0$ are the \emph{input frequencies}, and $\varphi \!:=\! \mathbf{b}^0$ are~the \emph{phase~shifts}.  
The final layer $\mathbf{L}$ is an affine~transformation.

The network's spectrum is defined by the input frequencies $\omega$ (see~\cref{sec:freqemb}).  
Small $\omega$ limit the representable spectrum, resulting in \textit{underfitting} and overly smooth reconstructions that fail to capture fine details.  
Conversely, a large $\omega$ allows sharper reconstructions but introduces \textit{overfitting} and high-frequency leakage into smooth regions, leading to noisy artifacts.
This trade-off arises because SIREN neurons are globally coupled; each parameter contributes to the entire domain, as we show in~\cref{lemma}.  

\vspace{0.1cm}
\noindent\textbf{Spatially-adaptive networks.}
To overcome these limitations, we propose to modulate SIREN activations with \textit{spatially adaptive masks} produced by a lightweight hash-grid MLP $\mathcal{M}_\phi$.  
This mechanism localizes frequency contributions across space by constraining each neuron's effective spatial support.  
Formally, we replace each sinusoidal layer $\mathbf{S}^i$ with its spatially modulated counterpart:
\begin{align}\label{sasnet-neurons}
   \widetilde{\mathbf{S}^i} = \mathcal{M}^i \odot \mathbf{S}^i,\text{ where } \mathcal{M}^i(\mathbf{x}) \in [0,1]^{n_i}
\end{align}
are the learned spatial masks and $\odot$ denotes the Hadamard product.  
Substituting the modulated neurons from~\eqref{sasnet-neurons} into the MLP definition in~\eqref{siren} gives rise to \textbf{SASNet}, a model designed to mitigate overfitting and improve the parameter efficiency of SIRENs via \textit{frequency localization}.  

By construction, SASNet consists of a frequency embedding layer (\cref{sec:freqemb}), spatially-adaptive masks (\cref{sec:smask}), and a joint training scheme that optimizes the masks via a multi-scale hash-grid MLP strategy.  
This design allows SASNet to selectively activate high-frequency components where needed while maintaining smoothness and stability across the signal domain.

\subsection{Frequency embedding layer}\label{sec:freqemb}

We show that the SIREN spectrum is fully determined by its \emph{input frequencies}.  
Following the trigonometric identity in~\cite{novello2024taming}, each sinusoidal neuron expands as a sum of sines whose frequencies are integer linear combinations of the input frequencies.  
Hence, the neuron modulation in~\eqref{sasnet-neurons} serves to spatially localize these generated frequency components.

Precisely, recall that an \( ij\)-\textit{neuron} \( h_j^i(\mathbf{x}) \) of a SIREN refers to the \( j \)-th neuron of the \( i \)-th layer \( \textbf{h}^i(\mathbf{x}) \):
\begin{align}\label{eq:neuron}
h_j^{i}(\mathbf{x}) = \sin\Biggl(\sum_{k=1}^n W_{jk}^{i} \underbrace{\sin(\mathbf{y}^{i-1}_k)}_{h^{i-1}_k(\mathbf{x})} + b_j^{i} \Biggl),
\end{align}  
where \( \mathbf{y}^{i-1} \) denotes \( \mathbf{h}^{i-1}(\mathbf{x}) \) before activation.
Note that weight \( W^{i}_{jk} \) determines the influence of neuron \( h_k^{i-1} \) in shaping the next layer. \citet{novello2024taming} shows that \( h_j^i(\mathbf{x}) \) expanded as follows:
\begin{theorem}\label{lemma}
    The neuron $h_j^{i}$ admits the following expansion:
    \begin{align}\label{e-neuron-expansion}
    h_j^{i}(\mathbf{x}) = \sum_{\mathbf{k}\in\Z^{n_i}}\alpha_{\mathbf{k}}\,\sin\Big( \dotprod{\mathbf{k}, \textbf{y}} + b_j^{i} \Big),
    \end{align}
    where $\alpha_{\mathbf{k}}=\prod_{l}J_{k_l}({W}_{jl}^{i})$ is the product of Bessel functions.
\end{theorem}
\cref{lemma} shows that the frequencies generated in the second layer \(\mathbf{h}^2(\mathbf{x})\) are integer linear combinations \(\langle \mathbf{k}, \omega \rangle\) of the input frequencies \(\omega\).  
Moreover, this result generalizes to deeper layers by inductively applying \cref{lemma} across the network depth.  
Hence, the input frequencies \(\omega\) determine the full network spectrum, given by \(\langle \mathbf{k}, \omega \rangle\).

\vspace{0.1cm}
\noindent\textbf{Input frequency initialization.} We initialize $\omega$ so that the generated frequencies $\langle \mathbf{k}, \omega \rangle$ cover a broad spectral range while preserving accurate low-frequency reconstruction. We define the network over periodic functions and ensure the samples span a complete period. Since such functions can be written as a Fourier series, \ie, sums of sine terms with integer frequencies, \cref{lemma} shows that SIRENs can be formulated in this form. Because real-world signals typically have larger low-frequency amplitudes, we sample most elements of $\omega$ from the low-frequency range $[-\mathcal{L}, \mathcal{L}]^{n_0}$ and the rest from a higher-frequency band. This defines a fixed \textit{frequency embedding layer}, while all other weights use the standard initialization scheme~\cite{sitzmann2020implicit}. Details are provided in the supplementary (Supp. Sec 3.1).

Unlike standard SIRENs, where frequencies evolve during training, our formulation keeps the spectrum fixed while learning only the amplitudes \(\alpha_{\mathbf{k}}\).  
This approach stabilizes optimization and accelerates convergence~\cite{novello2024taming}.

\subsection{Spatially-adaptive masks}\label{sec:smask}

\begin{figure}[h!]
\vspace{-0.2cm}
    \centering
    \includegraphics[width=0.9\columnwidth]{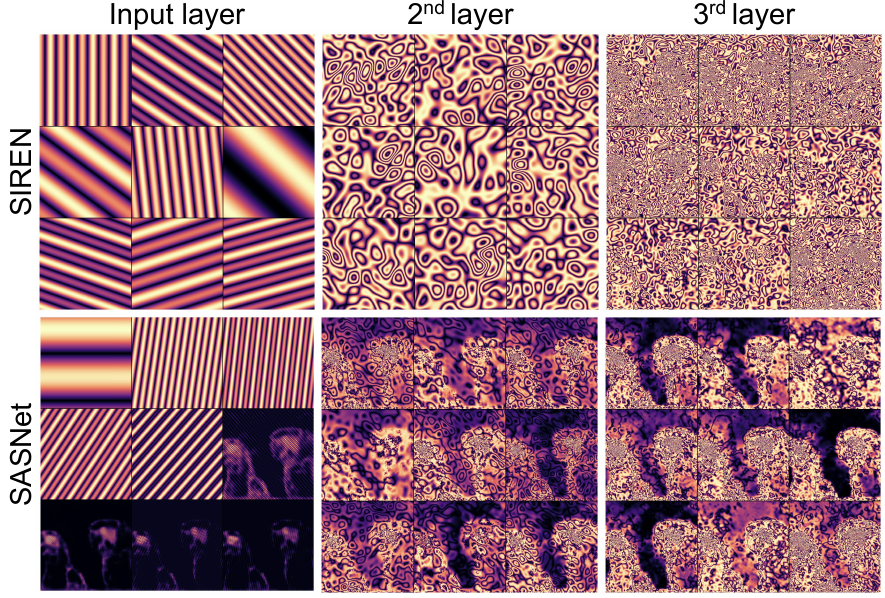}
    \vspace{-0.2cm}
    \caption{
\textbf{Neuron visualization}
for {SIREN} and {SASNet} on the Kodak “two macaws” image.  
Each column shows 9 neuron activation maps from the input, 2\textsuperscript{nd}, and 3\textsuperscript{rd} layers.  
While SIREN exhibits global activations and high-frequency interference, SASNet produces spatially localized and semantically aligned patterns.}\label{fig:influence}
\end{figure}

\cref{lemma} shows that the amplitudes $\alpha_{\mathbf{k}}$ of the frequencies generated by each hidden neuron $h^i_j(\mathbf{x})$ are independent of the spatial coordinate~$\mathbf{x}$.  
Consequently, high-frequency components may contribute \emph{globally} across the domain, including smooth regions, motivating the introduction of \emph{spatially adaptive masks} to localize these frequency contributions.
To enable spatial control, we multiply each neuron $h^i_j(\mathbf{x})$ by a mask $\mathcal{M}^i_j(\mathbf{x}) \in [0,1]$, modulating its influence over the domain.  
Thus, SIREN neurons are replaced by
$
\widetilde{h^i_j}(\mathbf{x})
    = h^i_j(\mathbf{x}) \, \mathcal{M}^i_j(\mathbf{x}),
$
and substituting into the expansion in~\eqref{e-neuron-expansion} yields the spatially localized form:

\vspace{-0.3cm}
\begin{align}
\widetilde{h^i_j}(\mathbf{x})
    = \sum_{\mathbf{k}\in\mathbb{Z}^{n_i}} 
     \alpha_{\mathbf{k}} \, \mathcal{M}^i_j(\mathbf{x})\,
      \sin\!\Big( \langle \mathbf{k}, \mathbf{y} \rangle + b_j^{i} \Big).
      \vspace{-0.2cm}
\end{align}
Thus, the proposed modulation makes the frequency amplitudes $\alpha_{\mathbf{k}} \, \mathcal{M}^i_j(\mathbf{x})$ {spatially dependent}, allowing high-frequency responses to be activated only where needed.  
\cref{fig:influence} shows neurons of SIREN and SASNet. Notice that the mask $\mathcal{M}^1(\mathbf{x})$ applied to the frequency embedding layer  
$\mathcal{M}^1(\mathbf{x}) \odot \sin(\omega \mathbf{x} + \varphi)$ directly controls which frequencies appear in each region, as observed in our experiments.

\vspace{0.1cm}
\noindent\textbf{Hash-grid network.}
We parametrize the spatially adaptive masks $\mathcal{M}_\phi$ using a multi-scale hash-grid network~\cite{muller2022instant}, which allows the masks to be jointly optimized with the SIREN parameters.  
To avoid the memory and computational cost of learning a different mask for every neuron $h^i_j(\mathbf{x})$, we partition the neurons of each layer $i$ into $n$ groups.  
Each group is assigned a single mask $\mathcal{M}^i_j(\mathbf{x})$, which is then \emph{broadcasted} to all neurons within that group, modulating their outputs.
Additionally, each input neuron $\sin(\omega_i \mathbf{x} + \varphi_i)$ is grouped according to its maximum frequency $\max(|\omega_i|)$.  
Specifically, neurons satisfying $\max(|\omega_i|) < \mathcal{L}$ are assigned to a low-frequency group, while the remaining neurons are partitioned into groups based on the magnitude of $\max(|\omega_i|)$.  
In contrast, neurons in the hidden layers are evenly divided into $n$ groups.

\vspace{0.1cm}
\noindent\textbf{Training.}  
To enforce sparsity, we sum a term to the loss $\mathscr{L}(\theta)$ that encourages the masks to take values close to zero:

\vspace{-0.3cm}
\begin{equation*}
\mathscr{L}(\theta) \;+\; \lambda\int \lvert \mathcal{M}_\phi(\mathbf{x}) \rvert_1\,\mathrm{d}\mathbf{x}.
\vspace{-0.1cm}
\end{equation*}
where $\lambda$ controls the strength of the regularization. Computed over sampled spatial coordinates $\mathbf{x}$, this extra term penalizes the mask values $\mathcal{M}^i_j(\mathbf{x})$, encouraging the usage of high-frequency components only when necessary.

\section{Experiments}\label{sec:exp}
We evaluate SASNet on three tasks: 2D image fitting (\cref{sec:imgfitting}), 3D volumetric data reconstruction (\cref{sec:volfitting}), and signed distance function (SDF) reconstruction (\cref{sec:sdf}). We also conduct an ablation study (\cref{sec:ablation}) to assess the contribution of each component in our proposed method.

\subsection{Image fitting}\label{sec:imgfitting}
We evaluate SASNet against state-of-the-art INRs with open-source implementations (summarized in~\cref{tab:div2k}) on the task of image fitting. Each $512^2$ RGB image of DIV2K~\cite{Agustsson_2017_CVPR_Workshops} is fitted by an INR and compared to the ground truth (GT) using the \textit{peak signal-to-noise ratio} (PSNR) and \textit{structural similarity index measure} (SSIM)~\cite{wang2004image}.

However, these metrics fail to fully capture noisy reconstruction artifacts caused by excessively high-frequency components, particularly in smooth regions. To detect such regions, we identify edges using Canny edge detector~\cite{canny1986computational} and dilate them with a disk kernel, resulting in a binary mask that separates edge and smooth areas, i.e., $\mathds{1}_{\text{edge}}$ and $\mathds{1}_{\text{smooth}}$.
To quantify noise, we compute the discrepancy in gradient norms between the input image and the model output in the smooth regions, resulting in the \textbf{noisiness} metric:
\begin{align*}
   \frac{1}{N_{\text{smooth}}} \sum_{i=1}^N \mathds{1}_{\text{smooth}}(\mathbf{x}_i)\,
   \Big|\,\lVert\nabla \mathscr{f}_{\text{gray}}(\mathbf{x}_i)\rVert_2 -
   \lVert\nabla f_{\text{gray}}(\mathbf{x}_i)\rVert_2 \Big|,
\end{align*}
where $\mathbf{x}_i$ are the image pixels, $N_{\text{smooth}}$ is the number of pixels in the smooth regions, and $\mathscr{f}_{\text{gray}}$ and $f_{\text{gray}}$ denote the ground-truth and predicted grayscale images. The INR gradient $\nabla f_{\text{gray}}$ is computed analytically using automatic differentiation, while $\nabla \mathscr{f}_{\text{gray}}$ is estimated using a second-order finite-difference method.
Finally, since the edge-region reconstruction quality is closely tied to image sharpness, we evaluate it using the \textbf{edge-region PSNR}:
\begin{equation*}
    \text{PSNR}_{\text{edge}} =
    -10 \log_{10}\big(\text{MSE}(\mathds{1}_{\text{edge}}\odot \mathscr{f},\,
    \mathds{1}_{\text{edge}}\odot f_\theta)\big).
\end{equation*}
In all image experiments, the compared methods are trained using their recommended hyperparameter settings.

\subsubsection{Quantitative comparisons}\label{sec:quant}
\cref{tab:div2k} compares SASNet with classical and state-of-the-art INRs on the image fitting task, showing that our method consistently achieves high reconstruction accuracy while maintaining very low noise. %
All models are trained for $5\text{K}$ epochs, except GaussianImage, which requires $50\text{K}$ epochs to converge.  
To ensure a fair comparison, all architectures are configured to have a similar number of parameters.
SASNet's spatial mask generation branch introduces only modest computational and GPU memory overhead, which is analyzed in the supplementary (Supp. Sec. 4).
\begin{table*}[tbp]
\centering
\small
\caption{\textbf{Quantitative evaluation of image fitting} on DIV2K~\cite{Agustsson_2017_CVPR_Workshops}. The \colorbox{red!25}{best} and \colorbox{orange!25}{runner-up} values are highlighted in color boxes. \textbf{Act func.} denotes the activation function, \textbf{FE} indicates the use of a frequency or positional embedding technique, and \textbf{BC} denotes an explicit control of the output signal's band limit, \textbf{Type} classifies INRs between ``Global", ``Hybrid", and ``Local" localization methods.}\label{tab:div2k}
\vspace{-0.2cm}
\begin{tabular}{l|cccc|cccl}
\toprule
\textbf{Methods} & \textbf{PSNR $\uparrow$} & \textbf{SSIM $\uparrow$} & \textbf{$\text{PSNR}_\text{edge} \uparrow$} & \textbf{Noisiness $\downarrow$} & \textbf{Act func.} & \textbf{FE}  & \textbf{BC} & \textbf{Type} \\
\midrule
SIREN~\cite{sitzmann2020implicit}            & 34.64                         & 0.959                       & 31.37                       & \cellcolor{orange!25}2.525      & $\sin$                &             &             & Global       \\
FFN~\cite{tancik2020fourier}                 & 31.92                         & 0.910                       & 30.20                       & 9.873                        & \texttt{ReLU}           & \checkmark  &             & Global       \\
SAPE~\cite{hertz2021sape}                    & 32.57                         & 0.932                       & 30.31                       & 3.128                        & \texttt{ReLU}           & \checkmark  & \checkmark  & Hybrid       \\
WIRE~\cite{saragadam2023wire}                & 28.76                         & 0.831                       & 25.73                       & 10.324                       & \texttt{wavelet}  &             &             & Local       \\
FINER~\cite{liu2024finer}                    & \cellcolor{orange!25}37.52    & 0.967                       & \cellcolor{orange!25}34.02  & 4.617                        & $\sin((1 \!+\! |x|)x)$    &             &             & Global       \\
GaussianImage~\cite{zhang2024gaussianimage}  & 37.38                         & \cellcolor{orange!25}0.973  & 33.25                       & -                                      & -                     &             &             & Local        \\
NeuRBF~\cite{chen2023neurbf}                 & 36.33                         & 0.959                       & 33.92                       & 6.975                        & \texttt{ReLU}           & \checkmark  &             & Hybrid     \\
\midrule
SASNet (Ours)                                         & \cellcolor{red!25}39.82       & \cellcolor{red!25}0.979     & \cellcolor{red!25}36.81     & \cellcolor{red!25}2.105   & $\sin$                & \checkmark  & \checkmark  & Hybrid   \\
\bottomrule
\end{tabular}
\vspace{-0.3cm}
\end{table*}

\pagebreak
SIREN and NeuRBF reduce noise but underfit fine details, leading to lower PSNR.  
FFN, WIRE, and SAPE show weaker accuracy, either due to limited frequency capacity or unstable local modeling.  
FINER provides sharper reconstructions but introduces substantial high-frequency noise.  
In contrast, {SASNet achieves the best overall performance across all metrics}.  
Its spatially adaptive masks avoid unnecessary high-frequency activations in smooth areas while enhancing detail reconstruction where needed, leading to the highest $\text{PSNR}_{\text{edge}}$ and the lowest noisiness scores.
Additional results are provided in the supplementary, where we further evaluate SASNet on medical CT slices and headshot images, showing that the method generalizes effectively across diverse image modalities (Supp. Sec. 5).

\begin{figure}[h!]
\vspace{-0.2cm}
    \centering
    \includegraphics[width=\columnwidth]{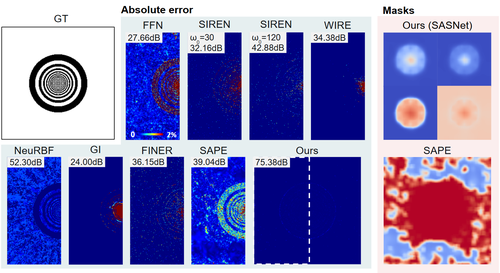}
    \vspace{-0.6cm}
    \caption{\textbf{Comparison of reconstruction error maps and learned masks on a toy image.}
The leftmost column shows the GT, and the absolute error maps for all methods.
SAPE and SASNet masks are shown on the right.
For SASNet, we visualize two frequency-embedding group masks (top) and two hidden-layer group masks (bottom), showing how the model activates neurons only around high-frequency regions.
GI refers to GaussianImage.}\label{fig:toy}
\vspace{-0.2cm}
\end{figure}

\subsubsection{Qualitative comparisons}\label{sec:qual}
To evaluate how effectively SASNet adapts to the spatial frequency distribution of a signal, we test all methods from \cref{tab:div2k} on a toy image containing sharp edges, flat regions, and a wide spectrum of frequencies (see GT in \cref{fig:toy}).
All models use an architecture with approximately $11\text{K}$ parameters.
\cref{fig:toy} shows absolute error maps and the masks learned by SAPE and SASNet.  
Most baselines—including SIREN, FFN, NeuRBF, SAPE, FINER, and WIRE—exhibit ringing artifacts and fail to capture the high-frequency structure in the central region, often producing noisy backgrounds.  
GaussianImage and WIRE better suppress noise but still struggle to fit sharp details.  
In contrast, SASNet delivers substantially cleaner reconstructions, achieving a {+23 dB PSNR gain} over the second-best method by accurately capturing edges while suppressing noise in flat regions.  
The learned masks further highlight the difference: SAPE's masks are noisy and spatially inconsistent, whereas SASNet produces well-localized, multi-frequency masks that activate only in high-frequency areas.

\begin{figure}[!h]
    \centering
    \includegraphics[width=\columnwidth]{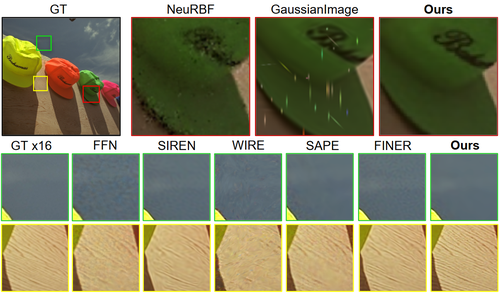}
    \vspace{-0.6cm}
    \caption{
\textbf{Qualitative comparison  of different methods under pixel shifting and $\times16$ super-resolution.  }
Red rectangles show the zoom-in view of pixel-shifting results: each model is queried at coordinates shifted by $1/8$ pixel along both axes to reveal interpolation artifacts.  
Green rectangles mark smooth background regions, while yellow rectangles highlight high-frequency regions used to assess super-resolution.  
The bottom rows show the $\times16$ reconstructions of these patches, showing that SASNet preserves sharp details in high-frequency areas while avoiding noise in smooth regions.}\label{fig:superres}
\vspace{-0.4cm}
\end{figure}

\cref{fig:superres} evaluates the generalization capability of SASNet in a Kodak image. 
We consider two scenarios: An image patch generated by a trained INR that takes as inputs the trained coordinates shifted by $1/8$ of pixel size, and zoom-in $\times16$ computed by evaluating the INRs between the training coordinates.

The top row in~\cref{fig:superres} (in red) compares NeuRBF, GaussianImage, and SASNet for the first scenario.
The noisy inference result of NeuRBF indicates severe overfitting, while the discrete nature and limited number of Gaussian kernels in GaussianImage lead to special rasterization artifacts. 
In the bottom row, we evaluate the super-resolution of several SOTA methods in a flat region (in green) and a texture-rich region (yellow squares).
SASNet is the only method capable of generalizing in high-frequency regions while preventing unwanted high-frequency noise in smooth areas.

\subsubsection{Model scalability}
\begin{figure}[!h]
    \centering
    \includegraphics[width=\linewidth]{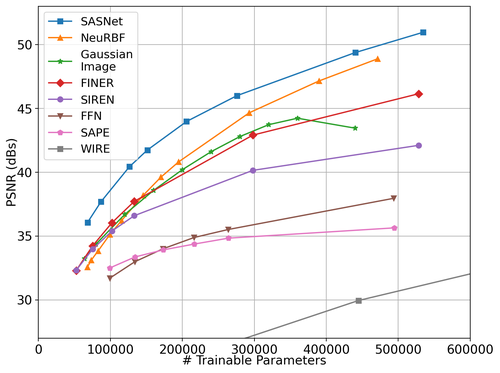}
    \vspace{-0.4cm}
    \caption{\textbf{Model scalability comparison}.
Image-fitting accuracy (PSNR) as a function of parameters. While all methods improve with increased capacity, SASNet shows the strongest scalability and maintains the best performance across all model sizes.}
    \label{fig:scalability}
    \vspace{-0.4cm}
\end{figure}

We evaluate the scalability of different INRs by varying the number of their trainable parameters and measuring the corresponding PSNR on the Kodak dataset~\cite{kodakdataset}.  
Model capacity is increased according to each method's architecture:  
hidden-layer widths for all INRs, number of Gaussians for GaussianImage, and for SASNet only the SIREN backbone is widened while the hash-grid and mask decoder remain fixed.
\cref{fig:scalability} shows that NeuRBF~\cite{chen2023neurbf}, FINER~\cite{liu2024finer}, and SASNet all benefit from increased capacity, whereas GaussianImage~\cite{zhang2024gaussianimage} saturates and eventually degrades beyond a certain threshold.  
SIREN, FFN, SAPE, and WIRE show weaker scaling behavior, with slower PSNR gains or early performance plateaus.
Across the entire parameter range, {SASNet consistently achieves the highest image-fitting accuracy}, highlighting that spatially adaptive masking provides a scalable and parameter-efficient mechanism for capturing complex frequency structures.

\subsection{Volumetric data representation}\label{sec:volfitting}

We further evaluate the generality of SASNet on {3D volumetric fields}, a setting where INRs must represent both smooth global structures and fine high-frequency details.  
Following the evaluation protocol of FINER~\cite{liu2024finer}, we use a smoke-density field 
$\rho:\mathbb{R}^3\!\to\!\mathbb{R}$ from the \textit{ScalarFlow} dataset~\cite{eckert2019scalarflow}, selecting 8 volumes from a $(100,178,100)$ grid.
We compare SASNet against five representative INRs:
PEMLP~\cite{mildenhall2021nerf}, SIREN~\cite{sitzmann2020implicit},  
Gauss~\cite{ramasinghe2022beyond},  
WIRE~\cite{saragadam2023wire},  
and FINER~\cite{liu2024finer}.  
All models are trained for $5$K epochs using the same architecture (3 layers with 256 neurons), except SASNet which uses 224 neurons per layer to keep the parameter count comparable.
\cref{tab:vol} reports quantitative results, showing that 
SASNet outperforms all baselines, achieving the highest PSNR and SSIM, with a noticeable margin over strong methods such as WIRE and FINER.
\begin{table}[h!]
\setlength\tabcolsep{3pt}
\small
\centering
\caption{\textbf{Quantitative comparison on volumetric data reconstruction} from ScalarFlow. The \colorbox{red!25}{best} and \colorbox{orange!25}{runner-up} results are highlighted.  
SASNet achieves the best reconstruction quality across all models, with the highest PSNR and SSIM.}
\vspace{-0.2cm}
\label{tab:vol}
\begin{tabular}{l|cccccc} 
\toprule
     & PEMLP  & Gauss  & SIREN  & FINER  & WIRE   & SASNet  \\
     \midrule
PSNR$\uparrow$ & 40.42  & 47.11  & 47.67  & 49.77  & \cellcolor{orange!25}53.06  & \cellcolor{red!25}55.92   \\
SSIM$\uparrow$ & 0.9434 & 0.9920 & 0.9948 & 0.9959 & \cellcolor{orange!25}0.9963 & \cellcolor{red!25}0.9973  \\
\bottomrule
\end{tabular}
\end{table}

\begin{figure}[h!]
    \centering
    \includegraphics[width=\linewidth]{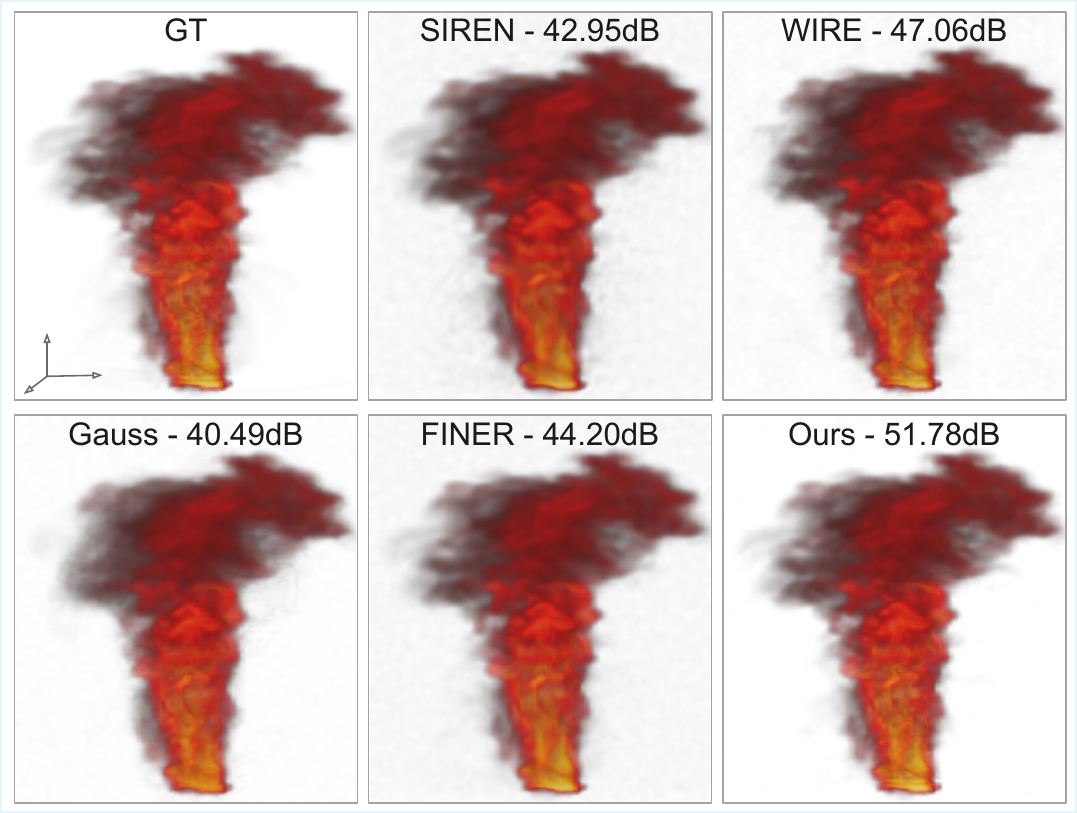}
    \vspace{-0.4cm}
    \caption{\textbf{Volumetric data reconstruction.}  
SASNet recovers fine-scale structures with sharper detail and reduces background noise more effectively than all baselines.}
    \label{fig:vol}
    \vspace{-0.4cm}
\end{figure}

Qualitative comparisons in \cref{fig:vol} further illustrate these improvements, showing that
SIREN, Gauss, and FINER reconstruct coarse structures but suffer from noise in empty regions due to global frequency leakage.  
WIRE preserves more high-freq details yet introduces artifacts in smooth areas.  
In contrast, SASNet produces the sharpest reconstruction while maintaining a clean background, capturing thin filaments and turbulent detail without amplifying noise.

\subsection{Signed distance functions}\label{sec:sdf}

Signed distance functions (SDFs) are fundamental implicit representations used in geometry processing, graphics, and simulation.  
Recent works show that INRs provide a powerful alternative to grid-based SDFs, preserving differentiability while improving compactness and fidelity~\cite{park2019deepsdf,novello2022exploring}.  
Thus, learning SDFs provides a robust benchmark for evaluating the stability and expressiveness of INR architectures.  

\vspace{0.1cm}
\noindent\textbf{Setup.}  
Following the open-source FINER pipeline~\cite{liu2024finer}, we evaluate four shapes with pre-sampled surface points and estimated normals.  
All baselines use the same architecture (3 layers × 256 neurons), while SASNet uses 224 neurons to keep the parameter count comparable.  
Training uses the coarse-to-fine SDF loss~\cite{yariv2021volume}, sampling $10$K points per iteration for $200$K iterations.  
For evaluation, we extract the zero-level set using Marching Cubes~\cite{lorensen1998marching} with $512^3$ grid and report Chamfer distance and IoU.

\begin{figure}[!h]
    \centering
    \includegraphics[width=\linewidth]{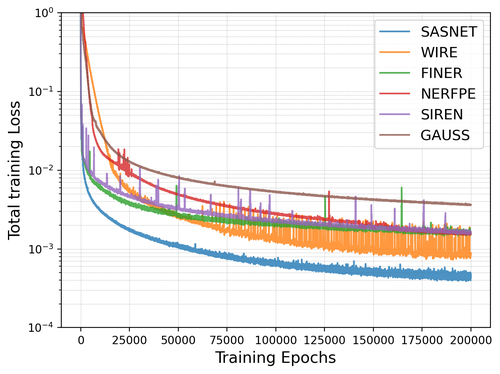}
    \vspace{-0.7cm}
\caption{\textbf{Training losses for the Armadillo SDF fitting task} on a log-scale, shown as moving averages with a window size of 35. SASNet reaches the final loss attained by the baselines within the first $\sim$35K iterations and achieves the lowest overall loss.}
    \label{fig:sdf_loss}
    \vspace{-0.3cm}
\end{figure}

As shown in~\cref{tab:sdf}, SASNet achieves the best reconstruction accuracy across all shapes and metrics.  
The fixed frequency embedding enables SASNet to represent high-frequency components of the SDF, capturing sharp curvature and intricate geometric features—particularly on complex shapes such as the \emph{Thai Statue}.  
At the same time, SASNet produces smooth, artifact-free fields in empty regions, avoiding the spurious surfaces that often arise from global INRs.
SASNet also converges significantly faster: as illustrated in~\cref{fig:sdf_loss}, it reaches FINER's final loss level within roughly {35K iterations}, and continues improving thereafter.  
This behavior confirms that the combination of spatially adaptive masking and fixed frequency embedding stabilizes the optimization and accelerates convergence.

\begin{figure}[t]
    \centering
    \includegraphics[height=\columnwidth]{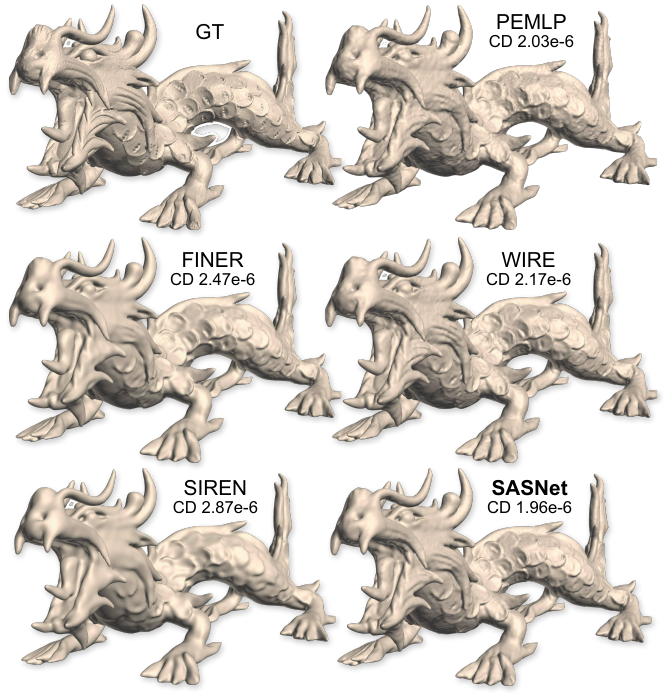}
    \vspace{-0.2cm}
    \caption{\textbf{Qualitative comparison on the SDF reconstruction task} for the \textit{Stanford Dragon}. SASNet produces the most accurate geometry, achieving the lowest Chamfer distance.}
    \label{fig:sdf_qual}\vspace{-0.5cm}
\end{figure}

\begin{figure}[!h]
    \centering
    \includegraphics[width=0.8\linewidth]{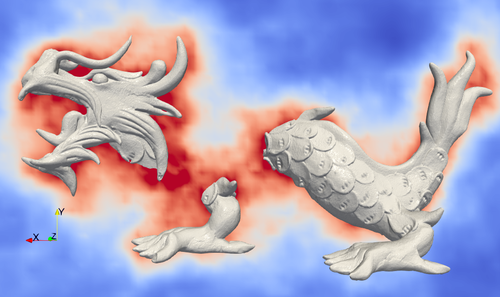}
    \vspace{-0.2cm}
    \caption{\textbf{Sliced 3D mask visualization} of SASNet's learned high-frequency spatial mask using a cool-warm color map.}
    \label{fig:sdfmask}\vspace{-0.4cm}
\end{figure}

\cref{fig:sdf_qual} shows reconstruction of the Stanford Dragon model, where SASNet recovers sharp geometric detail while maintaining clean, smooth regions, unlike other INRs, which exhibit noise, blurring, or missing structure. In addition, in~\cref{fig:sdfmask}, we visualize a cross-section of the learned mask for the highest-frequency band of the frequency embedding layer, showing that SASNet activates those neurons only near the object surface; full multi-view mask visualizations are provided in Supp. Sec. 5.2.

\begin{table}[h!]
\setlength\tabcolsep{4pt}
\scriptsize
\centering
\caption{\textbf{Quantitative comparison on SDF reconstruction.  }
The \colorbox{red!25}{best} and \colorbox{orange!25}{runner-up} results are highlighted.  
SASNet achieves the lowest Chamfer distance and highest IoU on all shapes, demonstrating superior geometric accuracy.}
\vspace{-0.2cm}
\label{tab:sdf}
\begin{tabular}{cl|ccccc}
\toprule
                                                  & Methods & Armadillo & Dragon    & Lucy      & Thai Statue & Avg.       \\ 
\midrule
\multirow{6}{*}{\rotatebox{90}{Chamfer$\downarrow$}} & PEMLP   & 3.465e-06 & \cellcolor{orange!25}2.039e-06 & \cellcolor{orange!25}2.387e-06 & 4.460e-06   & 3.088e-06  \\
                                                  & Gauss   & 7.283e-06 & 1.304e-05 & 1.178e-05 & 1.162e-05   & 1.093e-05  \\
                                                  & SIREN   & 3.599e-06 & 2.871e-06 & 2.405e-06 & 4.381e-06   & 3.314e-06  \\
                                                  & WIRE    & \cellcolor{red!25}3.274e-06 & 2.172e-06 & 2.419e-06 & \cellcolor{orange!25}3.296e-06   & \cellcolor{orange!25}2.790e-06  \\
                                                  & FINER   & 4.040e-06 & 2.477e-06 & 2.536e-06 & 4.201e-06   & 3.313e-06  \\
                                                  & SASNet  & \cellcolor{orange!25}3.368e-06 & \cellcolor{red!25}1.964e-06 & \cellcolor{red!25}1.927e-06 & \cellcolor{red!25}3.090e-06   & \cellcolor{red!25}2.587e-06  \\ 
\midrule
\multirow{6}{*}{\rotatebox{90}{IOU$\uparrow$}}       & PEMLP   & 0.9878    & \cellcolor{orange!25}0.9803    & 0.9677    & 0.9561      & 0.9730     \\
                                                  & Gauss   & 0.9806    & 0.9561    & 0.9581    & 0.9418      & 0.9592     \\
                                                  & SIREN   & 0.9872    & 0.9694    & \cellcolor{orange!25}0.9716    & 0.9561      & 0.9711     \\
                                                  & WIRE    & \cellcolor{red!25}0.9910    & 0.9747    & 0.9693    & \cellcolor{orange!25}0.9635      & \cellcolor{orange!25}0.9746     \\
                                                  & FINER   & 0.9837    & 0.9728    & 0.9671    & 0.9518      & 0.9689     \\
                                                  & SASNet  & \cellcolor{orange!25}0.9891    & \cellcolor{red!25}0.9809    & \cellcolor{red!25}0.9813    & \cellcolor{red!25}0.9654      & \cellcolor{red!25}0.9792     \\
\bottomrule
\end{tabular}
\end{table}

\subsection{Ablation studies}\label{sec:ablation}
Our procedure introduces two key components to enhance the performance of SIRENs: a frequency embedding layer and spatially adaptive masks for neuron groups. In~\cref{tab:ablation} we evaluate the contribution of each component to the baseline SIREN architecture.
Specifically, we consider adding the frequency embedding layer initialization (FEmbed.) and masking the input layer ($\textbf{h}^1$), hidden layers ($\textbf{h}^2, \textbf{h}^3$), or both. 

While initializing the frequency embedding layer provides the most significant improvement ($\sim$4 dB), additional masking strategies further improve accuracy. A qualitative comparison showing better reconstruction of high-frequency regions and background noise reduction is presented in the supplementary (Supp. Sec. 5.5).

\begin{table}[tbp]
\small
\caption{\textbf{Ablation study of SASNet for image fitting task} with different components enabled. \textbf{FEmbed.} stands for initializing the frequency embedding layer. $\textbf{Masks (h}^i\textbf{)}$ stands for masking neuron groups in layer $i$.}\label{tab:ablation}
\vspace{-0.2cm}
\centering
\begin{tabular}{c|ccc|cc}
\toprule
    \textbf{FEmbed.}   & \multicolumn{3}{c|}{\textbf{Masks}} & \textbf{PSNR$\uparrow$ } & \textbf{SSIM$\uparrow$} \\ 
              & $\textbf{h}^1$ & $\textbf{h}^2$ & $\textbf{h}^3$ &  \\ 
\midrule
    & & & & 36.00 & 0.921 \\
  \checkmark  & & & & 39.98 & 0.953 \\
  \checkmark  & \checkmark & & & 41.30 & 0.958 \\
  \checkmark  & \checkmark & \checkmark & & 41.44 & 0.959 \\
  \checkmark  & \checkmark &  & \checkmark & 41.75 & 0.963 \\
  \checkmark  &  & \checkmark &  & 41.08 & 0.962 \\
  \checkmark  &  &  & \checkmark & 40.56 & 0.960 \\
    & \checkmark & \checkmark & \checkmark & 37.10 & 0.932 \\
\midrule
  \checkmark  & \checkmark & \checkmark & \checkmark & \textbf{42.03} & \textbf{0.966} \\
\bottomrule
\end{tabular}
\end{table}
\vspace{-0.2cm}

\section{Conclusion}\label{sec:conclusion}
We presented {SASNet}, a novel \textit{Spatially-Adaptive Sinusoidal Network} that addresses key limitations of INRs. By combining a fixed frequency embedding layer with jointly optimized spatially-adaptive masks, SASNet improves frequency localization, enables explicit control over SIREN's spectral support, and suppresses frequency leakage. Experiments across various benchmarks show that SASNet consistently improves reconstruction quality while reducing noise artifacts, outperforming existing INR architectures in both quantitative and qualitative evaluations. Its gains become smaller when the model is heavily over-parameterized or the target signal is simple to fit. Future work includes reducing the overhead of spatial mask generation and exploring alternative initialization strategies.

\section*{Acknowledgment}
This work has been partially supported by the US National Science Foundation under grant numbers IIS-1910766 and IIS-2114451, and Google.

{
    \small
    \bibliographystyle{plainnat}
    \bibliography{main}
}

\end{document}


\maketitle
\tableofcontents

\section{Proof of Theorem 1}
In the main paper, we claim that the $j$-th neuron of the $i+1$-th layer $h_{j}^{i+1}$ admits the following expansion:
\begin{equation}\label{eq:thm1}
    h_{j}^{i+1}(\textbf{x})=\sum_{\textbf{k}\in\mathbb{Z}}\alpha\sin\left(\langle\textbf{k}, \textbf{y}\rangle+b_{j}^{i+1}\right)
\end{equation}
where $\textbf{y}=\textbf{W}^i\textbf{h}^{i}(\textbf{x}) + \textbf{b}^i$, and $\alpha=\prod_lJ_{k_l}\left(W_{jl}^{i+1}\right)$ are the products of Bessel functions of the first kind at the $j$-th row of the hidden matrix $\textbf{W}^{i+1}$.

Considering $h_{j}^{i+1}
\!=\!\sin\!\left(\!\sum_{k=1}^m W_{jk}^{i+1}\sin(y_k)+b^{i+1}_j\!\right)$, we now prove Eq.~\eqref{eq:thm1} by induction on $m$.
For $m=1$,

{\small
\begin{align*}
h_{j}^{i+1} = & \sin\left( W_{j1}^{i+1}\sin(y_1)+b^{i+1}_j\right) \\
= & \sin\left(W_{j1}^{i+1}\sin(y_1)\right)\cos(b^{i+1}_j)+\\ 
  & \qquad \cos\left(W_{j1}^{i+1}\sin(y_1)\right)\sin(b^{i+1}_j)
\end{align*}

\begin{align*}
= & \sum_{k\in\mathbb{Z} \text{ odd}}J_k(W_{j1}^{i+1})\sin(ky_1)\cos(b^{i+1}_j) + \\
  & \qquad \sum_{k\in\mathbb{Z} \text{  even}}J_k(W_{j1}^{i+1})\cos(ky_1)\sin(b^{i+1}_j) \\
= & \sum_{k\in\mathbb{Z} \text{  odd}}J_k(W_{j1}^{i+1})\sin(ky_1+b^{i+1}_j) + \\
  & \qquad \sum_{k\in\mathbb{Z} \text{  even}}J_k(W_{j1}^{i+1})\sin(ky_1+b^{i+1}_j) \\
= & \sum_{k\in\mathbb{Z}}\alpha\sin(ky_1+b^{i+1}_j).\\
\end{align*}
\vspace{-1.cm}
}

\noindent The first equality follows from the trigonometric identity $\sin(a+b)=\sin(a)\cos(b)+\cos(a)\sin(b)$, the second follows from standard identities for Bessel functions of the first kind (see~\cite[p. 361]{abramowitz1948handbook}). For the third, we use the properties $\sin(a)\cos(b)=\tfrac{\sin(a+b)+\sin(a-b)}{2}$ and $J_{-k}(a)=(-1)^kJ_k(a)$ to rewrite the summations. 

\noindent Now assume that Eq.~\eqref{eq:thm1} holds for $m-1$. Then,
{\small
\begin{align*}
h_{j}^{i+1}
\!= & \sin\!\left(\!\sum_{k=1}^m W_{jk}^{i+1}\sin(y_k)+b^{i+1}_j\!\right)\\
\!= &\sin\!\left(\!\sum_{k=1}^{m-1} W_{jk}^{i+1}\sin(y_k)+b^{i+1}_j\!\right)\cos\left(W_{jm}^{i+1}\sin(y_m)\right) + \\
& \quad \cos\!\left(\!\sum_{k=1}^{m-1} W_{jk}^{i+1}\sin(y_k)+b^{i+1}_j\!\right)\sin\left(W_{jm}^{i+1}\sin(y_m)\right) \\
\!= & \sum_{\substack{\textbf{k}\in\mathbb{Z}^{m-1}\\ l\in\mathbb{Z}\text{  } even}}\!\!\!\!\alpha_\textbf{k} J_{l}(W_{jm}^{i+1})\sin\left(\sum_{s=1}^{m-1}k_sy_s+b^{i+1}_j\right)\cos(ly_m) + \\
& \quad \sum_{\substack{\textbf{k}\in\mathbb{Z}^{m-1}\\ l\in\mathbb{Z}\text{  } odd}}\!\!\!\!\alpha_\textbf{k} J_{l}(W_{jm}^{i+1})\cos\left(\sum_{s=1}^{m-1}k_sy_s+b^{i+1}_j\right)\sin(ly_m) \\
\!= & \sum_{\textbf{k}\in\mathbb{Z}^{m}}\alpha\sin(\langle\textbf{k},\textbf{y}\rangle + b^{i+1}_j)
\end{align*}
Again, the second equality follows from the angle-sum identity for sine, and the third uses the induction hypothesis. This completes the proof.

\section{Contribution map}\label{append:contrib}
Analyzing the contribution of each neuron is an effective approach to understand a network's fitting capability and parameter redundancy. According to \citet{padmanabhan2024explaining}, for a neuron $h^{i}_{j}$, its contribution to each pixel coordinate $\textbf{x}$ is defined by a mapping from its weights $\textbf{W}^{i}_{j}$ to the neuron output scalar $h^i_j(\textbf{x})$, \ie, $\sum_l h^i_j(\textbf{x})\cdot \textbf{W}^{i+1}_{jl}$. Thus, the contribution map of $h^{i}_{j}$ is given by accumulating the pixel-wise contribution across the entire image. These maps illustrate the spatial representation focus of neurons\footnote{Publicly available at~\url{https://github.com/namithap10/xinc}}.

In the main paper, Fig. 3 presents contribution maps of nine neurons from SIREN~\cite{sitzmann2020implicit}, WIRE~\cite{saragadam2023wire}, and SASNet, representing different INR categories: \textit{Global}, \textit{Local}, and \textit{Hybrid} (see more examples in the Related work). The maps reveal that neurons in \textit{global} INRs tend to be redundant and \textit{``space-agnostic"}, whereas those in \textit{local} INRs exhibit sparser and more spatially localized contributions. Additionally, the global nature of INRs gives rise to model overfitting - neurons fitting high-frequency regions introduce noise into low-frequency areas. In such regions, \eg, image backgrounds, other neurons may need to counteract the effect of high-frequency noise by learning complementary signals to suppress these artifacts. This inefficiency in INRs is one of the main issues addressed in this paper.

\section{Implementation details}\label{append:impl}
\subsection{Frequency embedding layer initialization}

\begin{figure}[h]
    \centering
    \includegraphics[width=\columnwidth]{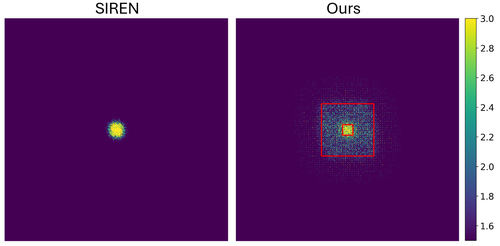}
    \caption{Comparing the frequency spectra of SIREN and our SASNet before training. Two red rectangles indicate band limits of low- and high-frequency embedding neurons.}
    \label{fig:initfft}
\end{figure}

As shown in~\cref{fig:initfft}, given a predefined band limit \( \mathcal{B} \) and a low-frequency range \( \mathcal{L} \), we split the frequency domain for selecting \( \omega \) into two concentric rectangular regions giving rise to the low- and high-frequency neurons.
Considering image-fitting as an example, low-frequency neuron multipliers \( \mathbf{k}_{low}\!\in\!\mathbb{Z}^2 \) are randomly sampled from \( [-\mathcal{L}, \mathcal{L}]^2 \). To ensure better coverage, high-frequency multipliers \( \mathbf{k}_{high} \) are sampled on an integer grid within \( [-\mathcal{B}, \mathcal{B}]^2 \), excluding the low-frequency region \( [-\mathcal{L}, \mathcal{L}]^2 \).  
To reduce redundancy, we constrain sampling to positive integers along the \( x \)-axis, leveraging a symmetry of the frequency domain, where \( (k_x, -k_y) \) is equivalent to \( (-k_x, k_y) \). By adjusting \( \mathcal{B} \) and balancing low- and high-frequency neurons, SASNet explicitly controls the output frequencies, improving high-frequency detail reconstruction. Choosing an appropriate \( \mathcal{B} \) mitigates overfitting and enhances the reconstruction of high-frequency details.

\subsection{Model configurations}
For the image fitting task, we follow each method's open-source implementation unless stated otherwise. For the sinusoidal baselines, \ie, FINER, SIREN, and SASNet, we increase $\omega_0$ from $30$ to $43$ because the images used here are of higher resolution than those in prior work, and this value gives a good trade-off between PSNR and reconstruction noisiness. All models are optimized with Adam using a fixed learning rate of $10^{-4}$, except as noted below.

\cref{tab:modelconfig} summarizes the task-specific SASNet hyperparameters used in our image fitting, volumetric fitting, and SDF reconstruction experiments. In all cases, the low-frequency mask is fixed to one, the remaining masks are initialized around $0.5$, and the hash-grid and decoder use a learning rate of $5\times10^{-4}$.
\begin{table}[h]
    \centering
    \caption{SASNet model configurations used in the supplementary experiments. For the SDF task, the band limit is $30$ for Armadillo and Dragon, and $40$ for Lucy and Thai Statue.}
    \label{tab:modelconfig}
    \footnotesize
    \begin{tabular}{lccc}
    \toprule
    Setting & Image & Volume & SDF \\
    \midrule
    Hardware & RTX A6000 & RTX A6000 & RTX A6000 \\
    Batch size & 262144 & 1780000 & 10000 \\
    Input dim & 2 & 3 & 3 \\
    $\omega_0$ & 43 & 30 & 30 \\
    \# trainable params & 128865 & 138181 & 130917 \\
    FEmb layer size & 256 & 256 & 256 \\
    Hidden layer size & 232 & 192 & 224 \\
    Band limit & 60 & 30 & 30 (40) \\
    Low-freq range & 12 & 6 & 6 \\
    Low-freq percentage & 0.5 & 0.5 & 0.5 \\
    \# high-freq bands & 4 & 4 & 4 \\
    \# neuron groups & 8 & 8 & 8 \\
    \midrule
    Hash-grid & & &  \\
    ~~feature dim & 2 & 2 & 2 \\
    ~~feature std & 0.01 & 0.01 & 0.01 \\
    ~~feature bias & 0 & 0 & 0 \\
    ~~codebook bitwidth & 10 & 11 & 10 \\
    ~~num levels & 10 & 12 & 10 \\
    Mask decoder & \multicolumn{3}{c}{ReLU-MLP with one hidden layer of size 48} \\
    \bottomrule
    \end{tabular}
\end{table}

\noindent Additional implementation details for the compared methods are listed below:
\begin{itemize}
    \item FINER~\cite{liu2024finer}: the bias scale is set to 0.
    \item FFN~\cite{tancik2020fourier} and SAPE~\cite{hertz2021sape}: we use the implementation in SAPE~\cite{hertz2021sape}\footnote{Publicly available at~\url{https://github.com/amirhertz/SAPE}.}. For each training step, instead of sampling coordinates, we feed all coordinates for supervision. The number of frequencies is set to 256 according to its performance on the Kodak dataset.
    \item WIRE~\cite{saragadam2023wire}: we use the implementation that each network layer uses a complex Gabor function (except the output layer). The learning rate is fixed to 5e-4. For the image-fitting task, the hyperparameter \texttt{omega} is set to 20, and \texttt{sigma} is set to 30, as suggested by its implementation. For the toy example, the hyperparameter omega is set to 4, and the sigma is set to 10 according to a simple hyperparameter search.
    \item GaussianImage~\cite{zhang2024gaussianimage}: without applying quantization, we fitted each image with 16640 Gaussians for 50K training epochs.
    \item NeuRBF~\cite{chen2023neurbf}: we tune the hyperparameters and set \texttt{log2\_hashmap\_size} to 14 and \texttt{n\_levels} to 10.
    \item SASNet: the full task-specific config. is summarized in~\cref{tab:modelconfig}.
\end{itemize}

For volumetric data and SDF fitting, we use the implementation of FINER~\cite{liu2024finer}\footnote{Publicly available at: \url{https://github.com/liuzhen0212/FINER}} and reuse its training and evaluation scripts on the four provided training point clouds. The corresponding SASNet hyperparameters for these two tasks are listed in~\cref{tab:modelconfig}.

\section{Computation performance and overhead}\label{append:overhead}
\begin{table}[h]
    \centering
\addtolength{\tabcolsep}{4pt}
\caption{Efficiency comparison on the 2D image fitting task. We report average step time (AST) and peak GPU memory usage (PMem) for models with comparable numbers of trainable parameters. Lower values indicate better efficiency.}
\label{tab:efficiency}
\begin{tabular}{l|cc}
\hline
   Model             & \color{red!60}{AST$\downarrow$} & \color{blue!60}{PMem$\downarrow$}  \\
\hline
{SIREN}  & \textbf{29,224}        & \textbf{2075.50}        \\
\hline
{MSIREN} & 95,438        & 6077.07        \\
{INCODE} & 59,907        & 3642.59        \\
{SASNet} & \ul{43,107}        & \ul{3031.60}   \\
\hline
\end{tabular}
\end{table}

\begin{figure}[tbp]
\centering
  \includegraphics[width=0.8\columnwidth]{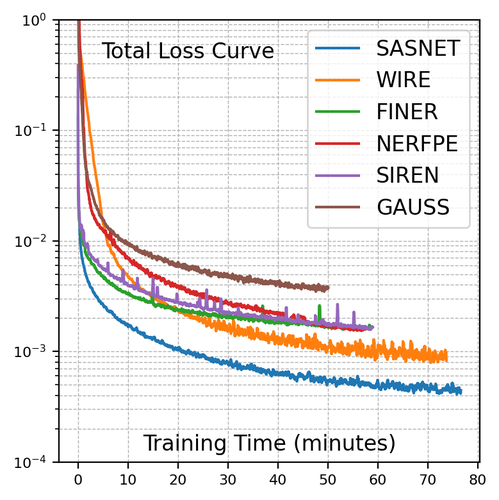}
  \caption{Training loss plotted against wall-clock time on the 2D image fitting task. SASNet reaches the same loss level substantially faster, highlighting its improved time-to-accuracy despite a modest per-step overhead.}
  \label{fig:wallclock_training_curve}
\end{figure}

We report {\color{red!60}{average step~time}} (AST,\textit{us}) and {\color{blue!60}{peak GPU mem usage}} (PMem, \texttt{MiB}) from PyTorch Profiler (averaged over 5 runs) with similar \#trainable params. 
For the image fitting task (Tab.~\ref{tab:efficiency}), we use batch size {262,144}; {INCODE}\citep{kazerouni2024incode} freezes the \texttt{feature\_extractor} and {MSIREN}\citep{mehta2021modulated} assumes precomputed latent codes and uses $32^2$ tiles with stride 16. For the 3D shape fitting task, we further profile SASNet and SIREN. SASNet increases AST from 17,557 to 22,175 and PMem from 98.0 to 118.2, indicating only a modest computational overhead.

This overhead mainly comes from the multi-scale hash-grid queries used to generate spatial masks. In practice, these gathers are often latency-limited because their memory accesses exhibit poor spatial locality and are therefore not well coalesced, making runtime sensitive to cache misses and memory-level parallelism. However, per-step efficiency alone does not fully reflect optimization behavior. When we redraw the training curve in the main figure as \cref{fig:wallclock_training_curve} with wall-clock time, SASNet reaches the same training loss in approximately 10 minutes, whereas SIREN requires 58 minutes. Overall, SASNet offers a favorable trade-off between model complexity and training efficiency: it incurs a moderate increase in runtime and memory usage, but converges substantially faster in practice. A more detailed runtime breakdown and optimization study is left for future work. In addition, the signal-conditioned modulation strategy used in INCODE suggests a possible direction for reducing this overhead further, namely predicting spatial masks with an INCODE-style module instead of generating them explicitly.

\section{More experimental results}\label{append:results}
\subsection{Influence of the hyperparameter $\omega_0$}
As shown in Fig. 1 of the main paper, SIREN~\cite{sitzmann2020implicit} is highly sensitive to the choice of the hyperparameter $\omega_0$, making its selection non-trivial. Smaller values lead to blurry reconstructions without introducing high-frequency noise, whereas higher values reduce spectral bias and produce sharper outputs, but resulting in increased noisiness in smooth regions.

In contrast, SASNet leverages spatially-adaptive masks to separate frequency band expressions, preventing high-frequency neuron outputs from propagating into smooth regions. As a result, the choice of $\omega_0$ has minimal impact on the final reconstruction accuracy, as illustrated in~\cref{fig:omega0}, where PSNR remains stable across different $\omega_0$ values.
\begin{figure}[h]
    \centering
    \includegraphics[width=1\linewidth]{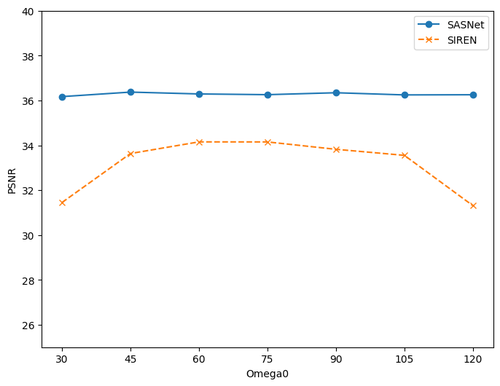}
    \caption{Sensitivity comparison between SIREN and our SASNet to the change of hyperparameter $\omega_0$.}
    \label{fig:omega0}
\end{figure}

\begin{table}[tbp]
\centering
\footnotesize
\caption{Quantitative evaluation on the Kodak~\cite{kodakdataset} dataset.}\label{tab:kodim}
\begin{tabular}{p{1.2cm}cccc}
\toprule
\textbf{Methods} & \textbf{PSNR $\uparrow$} & \textbf{SSIM $\uparrow$} & \textbf{$\text{PSNR}_\text{edge} \uparrow$} & \textbf{Noisiness $\downarrow$}\\
\midrule
SIREN~\cite{sitzmann2020implicit}            & 36.60                         & 0.951                       & 33.28                       & \cellcolor{orange!25}4.079      \\
FFN~\cite{tancik2020fourier}                 & 33.00                         & 0.868                       & 29.87                       & 10.353                       \\
SAPE~\cite{hertz2021sape}                    & 33.37                         & 0.872                       & 31.65                       & 5.373                         \\
WIRE~\cite{saragadam2023wire}                & 29.34                         & 0.804                       & 25.23                       & 15.375                       \\
FINER~\cite{liu2024finer}                    & \cellcolor{orange!25}37.72    & \cellcolor{orange!25}0.973                       & \cellcolor{orange!25}34.67  & 5.174                          \\
GI~\cite{zhang2024gaussianimage}  & 37.58                         & 0.971  & 32.59      & -                                     \\
NeuRBF~\cite{chen2023neurbf}                 & 37.51                         & 0.971                       & 34.78                       & 6.258                         \\
\midrule
SASNet                                         & \cellcolor{red!25}40.43       & \cellcolor{red!25}0.986     & \cellcolor{red!25}38.12     & \cellcolor{red!25}3.993    \\
\bottomrule
\end{tabular}
\end{table}

\subsection{Visualizations for the SDF}
\cref{fig:sdfterror} first visualizes the SDF error map for the Armadillo example. We compare a model equipped with the fixed frequency embedding but without spatial masks against the full SASNet model. Without spatial localization, the reconstruction error is distributed more broadly over the surface, especially around geometrically detailed regions, which indicates that globally available high-frequency components are not used selectively. In contrast, SASNet suppresses these artifacts and produces a substantially cleaner error distribution, showing that the learned masks help restrict high-frequency responses to the regions where they are most needed.

To further explain this behavior by the Dragon example in the main paper, \cref{fig:maskview} visualizes the learned spatial mask corresponding to the \textit{highest-frequency band} of the frequency embedding layer. The figure shows the target SDF isosurface for Dragon overlaid on the volumetric mask produced by the multi-scale hash-grid decoder. The mask values are rendered as a volumetric field using the same color encoding adopted for the 2D image masks. High mask responses concentrate near sharp geometric ridges, thin structures, and high-curvature surface regions, while remaining weak in smooth or empty space. This qualitative result confirms that SASNet allocates high-frequency capacity selectively, which helps explain the reduced error observed in \cref{fig:sdfterror}.

\begin{figure}[h]
    \centering
    \includegraphics[trim=0 15 0 10,clip,width=0.72\columnwidth]{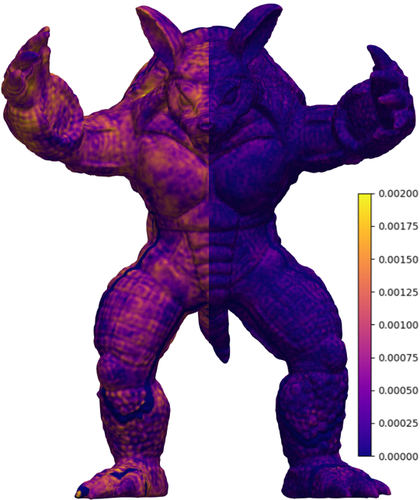}
    \caption{SDF error map for the Armadillo example. Left: a model with fixed frequency embedding but without spatial masks. Right: SASNet. The error map shows that spatially-adaptive masking suppresses widespread surface artifacts and localizes the remaining error to a smaller set of challenging regions.}
    \label{fig:sdfterror}
\end{figure}

\begin{figure*}
    \centering
    \includegraphics[width=0.9\linewidth]{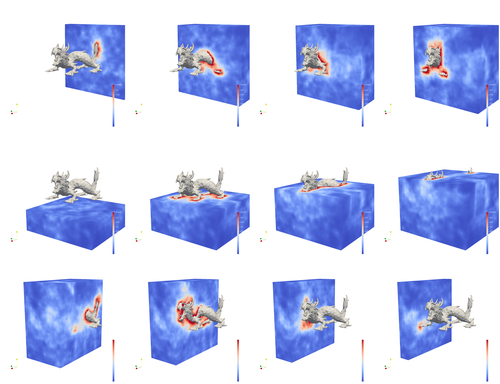}
    \caption{Learned 3D spatial mask for SASNet's highest-frequency band on the Dragon SDF. Across multiple viewpoints, high responses are concentrated near the object surface and around geometrically complex regions.}
    \label{fig:maskview}
\end{figure*}

\subsection{Additional experiments on CT medical images}
In addition to the two RGB image datasets used in the main paper, we further compare SASNet with FINER~\cite{liu2024finer} and SIREN~\cite{sitzmann2020implicit} on a publicly available CT medical image dataset\footnote{\url{https://www.kaggle.com/datasets/kmader/siim-medical-images}}, which contains 100 CT scan images from the TCIA cancer imaging archive~\cite{albertina2016cancer}. Each scan has a resolution of $512^2$ and consists of a single-channel grayscale image, normalized to $[0, 1]$. We conduct the image-fitting experiments using the same configuration as in the main paper. These results further demonstrate that SASNet generalizes well across modalities and resolutions.

\cref{tab:ctscan} presents the quantitative evaluations, showing that SASNet achieves higher fitting accuracy, particularly in preserving edge details. Notably, for single-channel images, the model configuration used in previous experiments is sufficiently large for all methods to fit the data without overfitting. Consequently, the noisiness across all methods remains low. This is further observable in~\cref{fig:ctscan}, where both the model outputs and their error maps are visualized.

\begin{table}[tbp]
\centering
\footnotesize
\caption{Quantitative evaluation on a medical image dataset. For each metric, $\uparrow$/$\downarrow$ indicates the numerical value is \textit{higher}/\textit{lower} the method performs better.}\label{tab:ctscan}
\begin{tabular}{lcccc}
\toprule
\textbf{Methods} & \textbf{PSNR $\uparrow$} & \textbf{SSIM $\uparrow$} & \textbf{$\text{PSNR}_\text{edge} \uparrow$} & \textbf{Noisiness $\downarrow$} \\
\midrule
SIREN~\cite{sitzmann2020implicit} & 47.07 & 0.98 & 42.94 & 3.482 \\
FINER~\cite{liu2024finer}         & 48.49 & 0.99 & 45.56 & 3.488 \\
\midrule
SASNet (Ours)                     & 50.21 & 0.99 & 49.58 & 3.681 \\
\bottomrule
\end{tabular}
\vspace{0.1cm}
\end{table}

\begin{figure*}[tbp]
    \centering
    \includegraphics[width=0.8\linewidth]{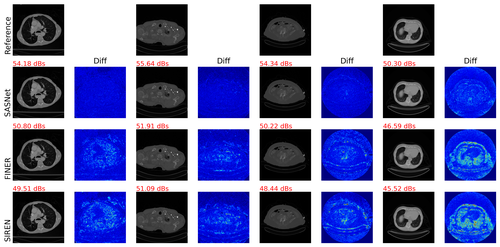}
    \caption{Medical CT scans' fitting results by SASNet, FINER, and SIREN. For each of the four examples, PSNR values are annotated on the upper left corner of the model outputs. Besides each model's output image, the \textit{Diff} column shows the error map with a \textit{jet} color encoding (up to 2\% absolute difference).}
    \label{fig:ctscan}\vspace{0.6cm}
\end{figure*}

\subsection{Additional experiments on FFHQ dataset}
The Flickr-Faces-HQ (FFHQ) dataset~\cite{karras2019style} contains a large collection of high-quality human face images at a resolution of $1024^2$. For this experiment, 100 images are randomly selected from the training image set. Similar to the medical image experiment, the same network configuration is applied to SIREN~\cite{sitzmann2020implicit}, FINER~\cite{liu2024finer}, and our SASNet.

\cref{tab:ffhq} presents the quantitative evaluations, while~\cref{fig:ffhq} provides a visual comparison among these methods. SASNet achieves higher fitting accuracy (as observed in the error maps), and this PSNR improvement does not result in an increased noisiness level, in contrast to FINER.
\begin{table}[tbp]
\centering
\footnotesize
\caption{Quantitative evaluation on the FFHQ dataset. For each metric, $\uparrow$/$\downarrow$ indicates the numerical value is \textit{higher}/\textit{lower} the method performs better.}\label{tab:ffhq}
\begin{tabular}{lcccc}
\toprule
\textbf{Methods} & \textbf{PSNR $\uparrow$} & \textbf{SSIM $\uparrow$} & \textbf{$\text{PSNR}_\text{edge} \uparrow$} & \textbf{Noisiness $\downarrow$} \\
\midrule
SIREN~\cite{sitzmann2020implicit} & 37.58 & 0.95 & 34.20 & 4.178 \\
FINER~\cite{liu2024finer}         & 38.16 & 0.95 & 35.09 & 4.975 \\
\midrule
SASNet (Ours)                     & 39.69 & 0.95 & 37.64 & 4.278 \\
\bottomrule
\end{tabular}
\end{table}

\begin{figure*}[tbp]
    \centering
    \includegraphics[width=0.8\linewidth]{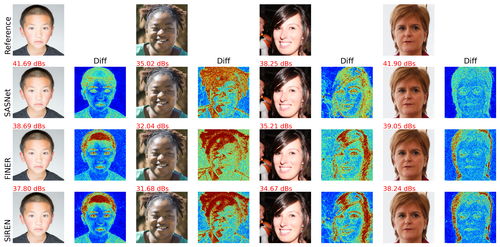}
    \caption{Fitting results of FFHQ human faces by SASNet, FINER, and SIREN. For each of the four examples, PSNR values are annotated on the upper left corner of model outputs. Besides each model's output image, the \textit{Diff} column shows the error map with a \textit{jet} color encoding (up to 2\% absolute difference).}
    \label{fig:ffhq}
\end{figure*}

\begin{figure*}[tbp]
    \centering
    \includegraphics[width=0.8\linewidth]{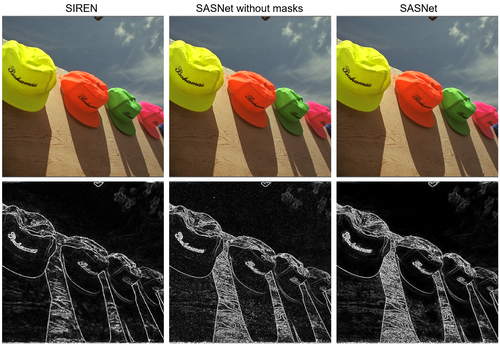}
    \caption{Visual comparison between SIREN, SASNet without masks, and full SASNet. First row shows the image reconstructed and the second row shows the corresponding gradient intensity obtained by network automatic differentiation (\textit{autograd} in PyTorch).}
    \label{fig:ablation}
\end{figure*}

\subsection{Additional ablation study results}\label{append:ablation}
In the main paper, we conduct an ablation study to evaluate the contribution of each module, namely the frequency embedding layer and spatially-adaptive masks at different layers. \cref{fig:ablation} presents a visual comparison among SIREN, SASNet with only frequency embedding (without masks), and the full SASNet. The results demonstrate that while the frequency embedding layer enhances fitting accuracy, particularly for sharper reconstructions, it also introduces additional noise in the image background. By applying spatially-adaptive masks, the quality of background reconstruction significantly improves, as indicated by the reduction in image gradient intensity.

\cref{fig:mask} visualizes the jointly optimized spatial masks of SASNet and SAPE for the same input image as in~\cref{fig:ablation}. The learned masks for different frequency bands effectively align with the spatial complexity of the image, enabling the model to focus its fitting capacity on complex, high-frequency regions while preventing resource waste and overfitting in simpler regions.

Additionally, a hidden mask is selected from each hidden layer, and contribution maps of the corresponding neuron groups are visualized. These results indicate that spatial masks allow different regions to encode distinct frequency bands, leading to lower network redundancy in SASNet compared to SIREN. In smooth and easy-to-fit regions, only a subset of neurons contributes to reconstruction, whereas in complex, texture-rich regions, most neurons participate in signal reconstruction, demonstrating the effectiveness of spatially-adaptive frequency control.

\section{Extreme cases}\label{append:extreme}
Under extreme cases, as shown in~\cref{fig:limitations}, INRs' performance can be limited for certain classes of image content. Both images primarily consist of high-frequency repeated patterns that span almost the entire domain, and with minimal pixel variation within each ``color strip". This structure is particularly well-suited for GaussianImage, as each Gaussian kernel can efficiently cover a larger portion of the domain with minimal computational effort. Conversely, INR methods suffer from spectral bias and an insufficient number of parameters to effectively capture such a broad frequency spectrum, leading to reduced fitting accuracy in these scenarios. These examples illustrate the inherent limitations of INRs under extreme global-frequency conditions.

\begin{figure*}[tbp]
    \centering
    \includegraphics[width=0.8\linewidth]{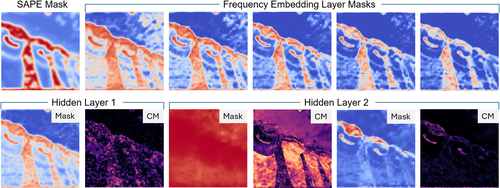}
    \caption{SAPE mask and our masks of the image-fitting task. \textbf{CM}: contribution map. The first hidden layer mask shows that this group has a global contribution, as also shown in the CM. However, some details on edges and texture regions can be missed, and therefore, the other group focuses only on these small regions.}
    \label{fig:mask}
\end{figure*}

\begin{figure*}
    \centering
    \includegraphics[width=\linewidth]{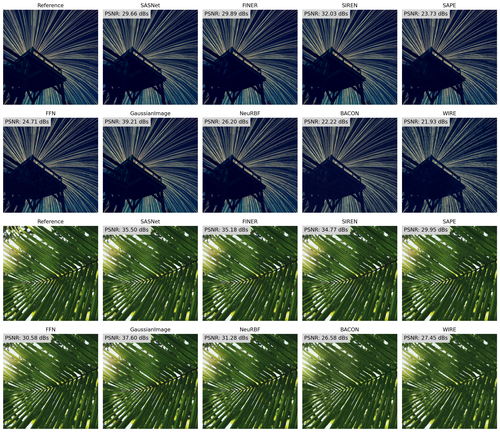}
    \caption{Example images from the DIV2K dataset showing when performances of INRs are limited. Both images are of complex signal components and higher frequencies than others, and contain hard edges covering almost the whole domain.}
    \label{fig:limitations}
\end{figure*}

{
    \small
    \bibliographystyle{plainnat}
    \bibliography{main}
}